\documentclass[conference]{IEEEtran}
\IEEEoverridecommandlockouts
\usepackage{cite}
\usepackage{amsmath,amssymb,amsfonts}
\usepackage{hyperref}
\usepackage[linesnumbered,ruled,titlenotnumbered]{algorithm2e} 
\usepackage{graphicx}
\usepackage{textcomp}
\usepackage{xcolor}
\usepackage{subcaption}
\usepackage{bm}
\def\BibTeX{{\rm B\kern-.05em{\sc i\kern-.025em b}\kern-.08em
    T\kern-.1667em\lower.7ex\hbox{E}\kern-.125emX}}
    
\def\c2p2{{\textsf{C2P2}}}
\def\C2P2{{\textsf{C2P2}}}

\newcommand{\cbase}[1]{\textsf{C2P2\ }\ }
\newcommand{\nop}[1]{{}}

\begin{document}

\title{C2P2: A Collective Cryptocurrency Up/Down Price Prediction Engine}
\author{\IEEEauthorblockN{Chongyang Bai\IEEEauthorrefmark{1}, 
Tommy White\IEEEauthorrefmark{1},
Linda Xiao\IEEEauthorrefmark{1}, 
V.S. Subrahmanian\IEEEauthorrefmark{1} and
Ziheng Zhou\IEEEauthorrefmark{2}
}
\IEEEauthorblockA{\textit{\IEEEauthorrefmark{1}Department of Computer Science, Dartmouth College, USA} \\
\{cy, tommy, lindaxs\}@cs.dartmouth.edu, vs@dartmouth.edu
}
\IEEEauthorblockA{\textit{\IEEEauthorrefmark{2}VeChain Foundation Limited, Singapore}\\
peter.zhou@vechain.com}

}

\maketitle
\begin{abstract}
We study the problem of predicting whether the price of the 21 most popular cryptocurrencies (according to \url{coinmarketcap.com}) will go up or down on day $d$, using data up to day $d-1$.  Our C2P2  algorithm is the first algorithm to consider the fact that the price of a cryptocurrency $c$ might depend not only on historical prices, sentiments, global stock indices, but also on the prices and predicted prices of other cryptocurrencies.  C2P2  therefore does not predict cryptocurrency prices one coin at a time --- rather it uses similarity metrics in conjunction with collective classification to compare multiple cryptocurrency features to jointly predict the cryptocurrency prices for all 21 coins considered.  We show that our C2P2 algorithm beats out a recent competing 2017 paper by margins varying from 5.1-83\% and another Bitcoin-specific prediction paper from 2018 by 16\%. In both cases, C2P2 is the winner on all cryptocurrencies considered. Moreover, we experimentally show that the use of similarity metrics within our C2P2 algorithm leads to a direct improvement for 20 out of 21 cryptocurrencies ranging from 0.4\% to 17.8\%. Without the similarity component, C2P2 still beats competitors on 20 out of 21 cryptocurrencies considered. We show that all these results are statistically significant via a Student's t-test with $p<10^{-5}$.
\end{abstract}


\section{Introduction}
        There are now over 1600 cryptocurrencies with a combined market capitalization exceeding \$100B\footnote{\url{https://en.wikipedia.org/wiki/List\_of\_cryptocurrencies}}.  The number of exchanges where these ``coins'' are traded is also growing.
    Despite initial suspicion about the use of cryptocurrencies for trading in illicit substances (e.g. the infamous Silk Road marketplace where drugs were allegedly traded for payments in Bitcoin\footnote{\url{https://blockexplorer.com/news/silk-road-timeline-bitcoin-drugs-dark-web/}, Dec 21 2018.}) and for managing payments to obtain keys to unlock files after ransomware attacks\footnote{https://cryptonews.com/exclusives/the-most-popular-cryptocurrencies-in-ransomware-attacks-1712.htm, May 4 2018.}, the use of cryptocurrencies for legitimate purposes has aggressively expanded in recent years. In 2014, \cite{glaser2014} found that many Bitcoin users view it as an asset rather than as a currency to exchange for goods, and that Bitcoin is used as an alternative investment to stocks and funds. Major companies (e.g. JP Morgan on Feb 19 2019) have even announced that they plan to launch their own cryptocurrencies\footnote{\url{https://www.thestreet.com/investing/bitcoin/jpmorgan-to-launch-first-united-states-bank-cryptocurrency-14866499}}. More interestingly, Revolut received a European banking license in December 2018, placing it more or less on par with normal banks, even though they trade in digital currency\footnote{\url{https://www.investinblockchain.com/cryptocurrency-banking-good/}}.
    Simply put, cryptocurrencies are now reaching the mainstream, and we can expect this trend to grow in coming years. 
    
    In this paper, we develop an algorithm and framework called Collective Cryptocurrency Price Prediction (\C2P2\ for short) to predict whether a cryptocurrency will move up or down in price one day ahead, i.e. \C2P2\ will predict at the end of day $d-1$, whether the price $P_{d}(c)$ of a particular cryptocurrency $c$ will move up (i.e. $P_{d}(c) > P_{d-1}(c)$) or move down (i.e. $P_d(c) < P_{d-1}(c)$) compared to the price $P_{d-1}(c)$ on day $d-1$. 
    
    We do \emph{not} focus on predicting the actual prices of cryptocurrencies in this paper because it is relatively easy to do so. To understand why, we implemented a simple and basic linear regression model to predict the next day's closing price using the prior day's closing price. 
    Notably, the Coefficient of Determination ($r^2$) values for all coins but Tether and NEM surpass 0.9, and Normalized Mean Square Error (NMSE) values range from 0.0001 to 0.0135. We chose to focus on the classification task of up/down price prediction because the baseline regression results were already impressive. 
    
    \C2P2\ involves several innovations on past work in cryptocurrency price prediction.  First, we comprehensively study the up/down daily opening, high, low, closing (OHLC) price movements of the 21 most popular cryptocurrencies\footnote{According to \url{coinmarketcap.com}. Note that this list is dynamic and the top cryptocurrencies are liable to change.} which is the most extensive study of up/down cryptocurrency price movements to date by number of cryptocurrencies considered. Because prices of cryptocurrencies can be affected by externalities (such as political events, law enforcement actions, natural disasters), and because the up/down movements of different cryptocurrencies may not be mutually independent, the \c2p2\ algorithm uses a novel iterative procedure: in the first iteration, we jointly predict the up/down movements of each of the 21 cryptocurrencies and save these as ``tentative'' predictions --- then in the next iteration, we use the tentative predictions from the previous iteration to learn a new predictive model and make a new set of tentative predictions --- and this process continues until either convergence occurs or until a specified stopping condition (number of iterations reaches a threshold). Second, \c2p2\ also captures pairwise similarities between the feature vectors of any two cryptocurrencies and leverages these similarities when making predictions. \c2p2\ also uses time-lagged variants of these features. We note that the \c2p2\ algorithm is the most important novel contribution of the paper. Third, we also identify the importance of different classes of features in predicting up/down cryptocurrency price movements. We show the importance of similarity features via an ablation study which shows that removing similarity features causes a significant drop in prediction results  (cf. Table \ref{tab:sim_contri} in the paper). Fourth, \C2P2\ incorporates a combination of feature selection and time delay lags --- we experimentally identified the best combination of features, time lags, and classification algorithms. Finally, unlike past work, our experiments use Reddit data, which serve to gauge the strength and polarity of public opinion on each cryptocurrency. Moreover, as cryptocurrency prices may be affected by stock markets, we also use stock index features from ten of the world's major stock markets, as well as gold, oil, and bond futures.

    We rigorously test the performance of our predictive models using the robust Area under a Receiver Operating Characteristic Curve (AUC) metric using a rolling window predictor to emulate real-world conditions.  Some past papers have used k-fold cross validation which is inappropriate for temporal data (which is our case).\footnote{K-fold cross validation takes a data set $D$ and randomly splits it into $k$ (usually $k=10$) disjoint subsets $D_1,\ldots,D_k$ called ``folds''. $k$ iterations are then performed. In the $i$'th iteration, the fold $D_i$ is used as the test set while the union of the remaining folds is the training set. Performance metrics for machine learning (e.g. accuracy, AUC, F1-score etc) are then aggregated across the performance across the $k$ folds. Because of the random split used in cross validation procedures, it is possible that the training fold ($\bigcup_{j\neq i} D_j$) contains data that applies to days in the future, which may then be used to predict up/down movements for days in the past that are present in the test fold $D_i$. This makes $k$-fold cross validation inappropriate for temporal data.}
    Instead, we use rolling window prediction in which training is done up to day $d-1$ and predictions are made for day $d$. Training is then done up to day $d$ and predictions are made for day $d+1$, and so forth. 
    
    
    Our experiments, conducted on 21 cryptocurrencies using six months of Reddit data from July 1st, 2018 to December 31st, 2018, used four months of data for training and then moved the training window one day at a time to predict up/down cryptocurrency price movements for the next day for the remaining two months.  We compared our results with those from a recent 2018 paper\cite{mcnally2018}, using market statistics and basic blockchain data available --- we also compared with results of a 2017 paper\cite{sin2017bitcoin}, though here we could only compare with Bitcoin as their paper used Bitcoin specific features. In our experiments, we make four types of predictions: Will day $d$'s high price exceed day $(d-1)$'s high price (High-High)? Will day $d$'s low price exceed day $(d-1)$'s low price (Low-Low)? Will day $d$'s closing price exceed day $(d-1)$'s (Close-Close)? And, finally, will day $d$'s opening price exceed day $(d-1)$'s (Open-Open)? Our experiments show that: 
    
    \begin{enumerate}
        \item \C2P2\ beats \cite{mcnally2018}'s method on all 21 cryptocurrencies by a margin of 5.1--83\% (depending on which of the four prediction types is considered) which is substantial and is statistically significant when comparing model performances across all 21 cryptocurrencies ($p < 10^{-5}$).
        \item \C2P2\ beats \cite{sin2017bitcoin} on Bitcoin data by 16\% for Close-Close task, which is also statistically significant ($p < 10^{-5}$). 
        \item The AUCs generated by \C2P2\ range from 0.609 to 0.8. Moreover, over half of the 21 cryptocurrencies are predicted with AUCs exceeding 0.7.
        \end{enumerate}
      
The results suggest that \C2P2\ is robust and offers good performance in predicting up/down movements for a variety of cryptocurrencies, outperforming recent baselines.
    
\section{Data Description}\label{sec:data}
    Our experiments and data apply to the second half of 2018 (from July 1 2018 to December 31 2018). We  
    gathered data from the following categories to use in our model.
    
    \emph{Historical market data}\footnote{Market data included daily OHLC prices.}  for the 21 cryptocurrencies studied was obtained from \url{https://www.binance.com/}. 
    
    \emph{Global economic metrics} aimed at capturing the fluctuations and volatility of the global economy, including ten well-known national stock indices\footnote{These ten national stock indices, alphabetically, were: ASX 200, CAC 40, DAX, DJIA, FTSE 100, IBEX, MOEX, Nikkei 225, BSE SENSEX, S\& P 500, and the SSE Composite Index.} as well as six indices related to commodities and bonds, were obtained from Yahoo Finance and Quandl. In total, we acquired 88 features from these indices for each day.
    
    \emph{Reddit data.} We used Reddit's PRAW API\footnote{\url{https://praw.readthedocs.io/en/latest/}} in combination with the PushShift API\footnote{\url{https://pushshift.io/}} in order to obtain posts (also known as submissions) from 23 relevant discussion communities on Reddit during the July 1, 2018 to December 31, 2018 time frame. In all, we gathered 225,869 submissions from 43,139 submitters --- additionally, we gathered 1,624,674 comments (on the submissions) from a total of 101,564 authors.

\section{Feature Extraction}\label{sec:arch}

    For each cryptocurrency $c$ and day $d$, we extract many features in order to create a feature vector $\bm{f}_{c,d}$ (cf. Algorithm \ref{alg}). The features fall into the following broad categories. 
    
    \emph{Reddit statistics.} To allow us to compare the daily distributions of basic reddit statistics, we calculate histograms across the comment counts and submission scores for all Reddit submissions made on each day. This process results in 22 features for each cryptocurrency on each day.
    
    \emph{Reddit psychological variables.}  We use the Linguistic Inquiry and Word Count (LIWC) framework \cite{pennebaker2015liwc} to calculate all LIWC features on the title text, submission text, and comment text for the Reddit submissions. Developed by a psychologist, LIWC produces features capturing psychological processes (e.g. positive emotion, anxiety, anger), cognitive processes (e.g. certainty, doubt), perceptual processes (e.g. see, feel, hear), drives (e.g. affiliation, power, reward, risk), and more. In total, LIWC provides us with 279 features for each cryptocurrency on each day.
    
    \emph{Reddit sentiment variables.} We use the Google Cloud Natural Language API\footnote{\url{https://cloud.google.com/natural-language/}} (Cloud NL) to run high-quality sentiment analysis on the title text, submission text, and comment text for all Reddit submissions made on each day. The API provides a sentiment polarity and magnitude for each sentence in our text data, in addition to an overall polarity and magnitude. For each of the text categories (title, submission, comment), we use the sentence results provided by Cloud NL to generate histograms across sentiment polarity and magnitude on each day. After this process, we end up with 66 sentiment features for each cryptocurrency on each day.

    \begin{figure}
        \centering
        \includegraphics[width=\columnwidth]{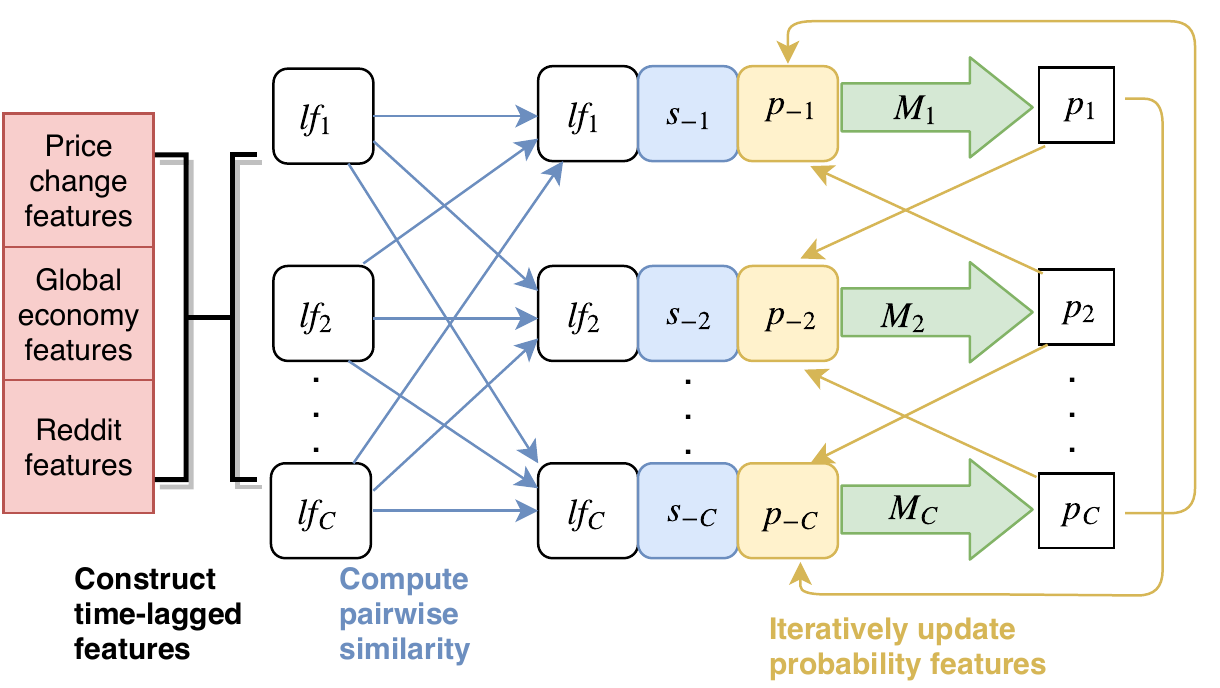}
        \caption{\c2p2 Architecture. To predict the price going up/down for day $d$, first, three sets of features are extracted before day $d$. Second, we build time-lagged feature vector $\bm{lf}_c$ for each cryptocurrency $c$, and compute pairwise similarities between time-lagged features of coins, then concatenate similarity of each \emph{other} coin with $c$ to $s_{-c}$ (excluding c with itself). We construct probability features for $c$ by concatenating all probabilities excluding $c$. Finally we iteratively make predictions by model $M_c$ and update probability features.}
        \label{fig:model_arch}
    \end{figure}

\section{\c2p2\ Algorithm}\label{sec:algo}

    We build our model on the idea of collective classification 
    \cite{kong2012meta,taskar2004link} where, instead of
    predicting the price of each of the 21 cryptocurrencies individually, we try to predict them simultaneously. However, our collective classification algorithm is novel in many respects --- first through the use of similarity metrics to compute pairwise similarities between the feature vectors of every pair of cryptocurrencies, and second, through the use of probabilities of prices going up rather than raw predictions (up vs. down).
    
    
    The \cbase\ algorithm is shown in detail as Algorithm~\ref{alg} and is visualized in Figure \ref{fig:model_arch}.  Informally speaking, the algorithm works as follows to predict whether the prices of cryptocurrencies will go up on day $d$.
    
    \begin{enumerate}
        \item The algorithm takes a set of $C$ cryptocurrencies as input - assume that the cryptocurrencies are therefore referenced as $1,\ldots,C$. The input features on or before day $d-1$ were described in Section \ref{sec:arch} and shown on the left red blocks in Figure \ref{fig:model_arch}. The algorithm additionally takes a lag $L$ as input saying that the last $L$ days of data (i.e. days $d-L,\ldots, d-1$) should be used for classification. For each considered cryptocurrency $c$, the algorithm uses a pre-trained classifier $M_c$ (we will describe the training stage shortly). Finally, the algorithm takes a similarity function $S$ to compare the feature vectors of any two cryptocurrencies, and a threshold $\epsilon$ to determine when the \c2p2 Algorithm converges.
        \item In Line~\ref{lin:init-prob}, \c2p2\ assumes that each cryptocurrency $c$ has a probability $p_{c,d}$ of going up (compared to the previous day) on day $d$. This probability is initially chosen uniformly at random, and a probability vector $\bm{p}_d$ describes the estimated probabilities on day $d$.
        \item Lines~\ref{lin:start-lf}--\ref{lin:end-lf} construct a time-lagged feature vector by considering the previous $L$ days. $L$ is an input to the algorithm and the choice of $L$ suggests that we think an up/down movement of the price on day $d$ depends only on the features of cryptocurrencies during the previous $L$ days, i.e. day $d-L$ through day $d-1$.  For each cryptocurrency $c$, we now create a \emph{time-lagged feature vector} $\bm{lf}_{c,d}$ of the concatenated feature vectors $\bm{f}_{c,d-L},\ldots,\bm{f}_{d-1}$.
        \item Lines~\ref{lin:start-sim}--\ref{lin:end-sim} look at each cryptocurrency $c$ and  compare the similarity of the time-lagged feature vector $\bm{lf}_{c,d}$ with the time-lagged feature vector of each \emph{other} cryptocurrency using a similarity function $S$. The resulting pairwise similarities are then concatenated together to create $\bm{s}_{-c,d}$. We consider five similarity functions: Euclidean, Cosine, Manhattan, Pearson Correlation Coefficient (PCC), and Spearman Correlation Coefficient (SCC). The idea in this step is that cryptocurrencies with similar time-lagged feature vectors should move up or down in a similar manner over time.
        \item The bulk of the classification task occurs within the outer loop shown in Lines~\ref{lin:out-loop-start}--\ref{lin:out-loop-end}.
        \begin{enumerate}
            \item In the $i$th iteration of this loop, we first create a copy (Line~\ref{lin:p-prime}) of the probability vector $\bm{p}_d$ for the day $d$ we are trying to predict. This is what the loop tries to update every time it is executed.
            \item Next, Line~\ref{lin:set-p} creates a vector of all cryptocurrency probabilities (i.e. the probability of going up) for day $d$ with the exception of currency $c$.
            \item The inner for loop (Lines~\ref{lin:for-loop-start}--\ref{lin:for-loop-end}) takes \emph{independently} a pre-trained model $M_c$ and predicts the probability of price going up/down for every cryptocurrency in Line \ref{lin:update-probs} --- but to do so, it uses as additional features, the predicted probabilities for each \emph{other} cryptocurrency at last prediction and the similarities with each other cryptocurrency, thus embodying the collective spirit.
            \item Finally, Line~\ref{lin:update-probs} updates the probability vector.
            \item The algorithm terminates either when the probability vectors generated in two successive iterations are very close to each other or when the number of iterations exceeds a threshold and the algorithm eventually terminates in Line~\ref{lin:answer} by returning the final probability vector. 
        \end{enumerate}
    \end{enumerate}
    
    {
    \begin{algorithm}[t!]
    \small
        \caption{\label{alg}}
        \SetKwRepeat{Do}{do}{while}
    	\SetKwInOut{Input}{Input}\SetKwInOut{Output}{Output}
        \Input{Features $\bm{f}_{c,d-L},\ldots \bm{f}_{c,d-1} $, learned classifier $M_c$ $\forall c \in [1, \ldots C]$,
        similarity function $S(\cdot, \cdot)$, lag $L$,
        convergence threshold $\epsilon$, maximum iteration number $I$} 
        \Output{$\bm{p}_{d}=(p_{1,d},\ldots p_{C,d})$. Predicted probabilities of prices going up for all C cryptocurrencies on day $d$}
        \BlankLine
        \tcc{sampling from uniform distribution}
        $\bm{p}_d =(p_{1,d}, \ldots p_{C,d}) \sim U(0,1) $ \label{lin:init-prob}\\
        \For{$c \in [1, \ldots C]$} { \label{lin:start-lf}
        	\tcc{construct time-lagged features $\bm{lf}$}
        	$\bm{lf}_{c,d} = (\bm{f}_{c,d-L},\bm{f}_{c,d-L+1},\ldots \bm{f}_{c,d-1})$\\
        	} \label{lin:end-lf}
        \tcc{compute pairwise similarity of $\bm{lf}$} \label{lin:start-sim}
        \For{$c \in [1, \ldots C]$}{
            $\bm{s}_{-c,d} = \underset{\forall i \not= c}{\text{concat}}(S(\bm{lf}_{i,d}, \bm{lf}_{c,d}))$\\
            
            } \label{lin:end-sim}
        	$iter = 0$ \\
            \Do{$\|\bm{p}_d - \bm{p}'_d\|_2 > \epsilon$ or $iter < I$}{ \label{lin:out-loop-start}
                $iter = iter + 1$\\
                $\bm{p}'_d = \bm{p}_d$ \label{lin:p-prime}\\
                Set $ \bm{p}_{-c,d} = \underset{\forall i \not= c}{\text{concat}}(p_{i,d})$ for each $c\in[1,\ldots,C]$ \label{lin:set-p}\\
                \For{$c \in [1, \ldots C]$}{ \label{lin:for-loop-start}
                    
                    \tcc{model $M_c$ predicts for coin $c$ by concatenating 3 sets of features}
                    $p_{c,d} = M_c (\bm{lf}_{c,d}, \bm{s}_{-c,d}, \bm{p}_{-c,d})$\label{lin:update-probs}
                } \label{lin:for-loop-end}
            }\label{lin:out-loop-end}
        \BlankLine
    	\Return{$\bm{p}_t$}\label{lin:answer}
    \end{algorithm}
    }
    
    \emph{Training stage of \c2p2.}
    We first use the training set (i.e. data from days $i\in \{1,\ldots,d-1\}$) to 
    construct the feature vectors $\bm{lf}_{c,i}, \bm{s}_{-c,i}$ as specified in Lines \ref{lin:init-prob}-\ref{lin:end-sim} of the \c2p2\ algorithm. For each $i \in [1, \ldots d-1]$ and each cryptocurrency $c$, we also have the corresponding labels $y_{c,i}$ of whether prices go up/down from the previous day. Second, we augment the feature vectors for each cryptocurrency-day pair $(c,i)$ 
    with the vectors $\bm{p}_{-c,i}$ so that in the training data, each feature vector also has the predicted probabilities (by \c2p2 ) as a feature. Each model $M_c$ is then trained iteratively until the training probability vectors ($\bm{p}_{1}$,\ldots $\bm{p}_{d-1}$) converge or the maximum  number of iterations $I$ is reached. This process is the same as Line \ref{lin:out-loop-start}--\ref{lin:out-loop-end} in the \c2p2 algorithm except that in  Line \ref{lin:update-probs} we train $M_c$ instead. Finally, models $M_1,\ldots M_C$ are saved.
    
    \nop{
    By doing this, algorithm \c2p2\ returns a probability vector $(p_1,\ldots,p_C)$ in Line~\ref{lin:answer}.  This probability feature vector, where $p_i$ is the probability (after learning from the training data) that cryptocurrency $i$'s price will increase on the next day, is then used for the actual prediction.
    }
    
    \emph{Example.} As an illustration of Figure \ref{fig:model_arch} and Algorithm \ref{alg}, consider the case when we have just two cryptocurrencies, Bitcoin and Ethereum ($C=2$) and suppose the lag considered is $L=2$. For each day, we first build time-lagged features $\bm{lf}_{1}$ for Bitcoin by concatenating the features (red blocks on the left of Figure \ref{fig:model_arch}) of the previous two days. We construct $\bm{lf}_{2}$ for Ethereum in the same way. Second we compute the similarity $s$ (Euclidean distance for example) of $\bm{lf}_{1},\bm{lf}_{2}$. In this case, the similarity would be the same, i.e. $s_{-1}=s_{-2}=s$ for some $s$. Third, we initialize probabilities $p_1,p_2$ and construct $\bm{p}_{-1}=p_2, \bm{p}_{-2}=p_1$. Then, $\bm{p}_{-1}, \bm{p}_{-2}$ are augmented with ($\bm{lf}_1,\bm{s}_{-1}$), ($\bm{lf}_2, \bm{s}_{-2}$) to get the feature vectors for models $M_1,M_2$. To predict whether their prices will go up/down on day 10, we train models $M_1,M_2$ using feature vectors and up/down labels from days 1 to 9.  During the training stage, we iteratively train $M_1,M_2$, update $p_1,p_2$, and update feature vectors for all days \emph{but only on the training data}. When the probabilities converge or the maximum number of iterations is reached, we save $M_1,M_2$. When predicting, we construct feature vectors on day 10 in the same way, then use learned $M_1,M_2$ to iteratively predict $p_1,p_2$ and update feature vectors. Finally, $p_1,p_2$ are returned when they converge or the number of iterations reaches a limit.
    
\section{Experimental Results}
    \subsection{Experiment setup}
        In our experiments, we vary the lag value $L$ from 1 to 30. We set the convergence threshold $\epsilon=10^{-3}$ and maximum iterations $I$ to $10$. Since \c2p2\ can use any basic classifier, we conduct our experiments with five: Logistic Regression (LR), Random Forest (RF), K Nearest Neighbors (KNN), Linear SVM (L-SVM) and Gaussian Naive Bayes (NB).
        
    \subsection{Baselines}
        We compared \c2p2 with two state-of-the-art baselines (one each from 2017 and 2018) to gauge to performance of our algorithm.
        
        The first baseline is \cite{sin2017bitcoin}, which used a genetic algorithm to choose an ensemble of Neural Networks to predict Bitcoin price. Because \cite{sin2017bitcoin} used data from 10 different Bitcoin exchanges in order to predict whether the next day's Bitcoin price would go up/down, and since these same data were not available for other coins, we were only able to compare \c2p2 against their method on Bitcoin data. The data used is from \url{Bitcoinity.org} and \url{Blockchain.info}.
        
        The second baseline is \cite{mcnally2018}, which found that LSTM RNNs worked best in Bitcoin price prediction. The authors used historical market statistics (such as price, volume, and market capitalization) in addition to blockchain features (such as block size, cost per transaction and hash rate) as input. We already had the market statistics for all of the cryptocurrencies, and we were able to collect blockchain data for all cryptocurrencies except IOTA, Maker, Ontology, and VeChain from \url{https://coinmetrics.io/}. We still compared C2P2 to \cite{mcnally2018} for these coins, but simply did not use blockchain data for them.
        
    \subsection{Feature Selection} With lag $L$, \c2p2 uses $455\times L+120$ features ($88L$ economic features, $366L$ Reddit features, $L$ history price features, $100$ similarity features, and $20$ probability features). 
    To remove redundant features and avoid overfitting, we perform two kinds of feature selections on the \emph{training set}. The first method is Principal Component Analysis (PCA), from which we get 120 features (same as the number of training examples). The second method chooses the 512 features with the highest F-values using Analysis of Variance (ANOVA) techniques. 
    We report the best results among the two feature selection methods in all experiment results.
    
    \subsection{Experiment 1: Comparison of \c2p2 and baselines}
    In this experiment, we compare \c2p2 with \cite{mcnally2018} for all 21 cryptocurrencies and compare with \cite{sin2017bitcoin} for Bitcoin (as only Bitcoin-related features were used in \cite{sin2017bitcoin}). 
    
    We define the \emph{lift} of an algorithm $A$ w.r.t. to a baseline method $B$ to be $Lift=\frac{AUC(A)}{AUC(B)}$. A lift that is greater than one suggests that algorithm $A$ outperforms algorithm $B$, while a lift less than one suggests the opposite. 
    
   
    Table \ref{table:baseline_cc} shows \c2p2's best results for next day closing price up/down prediction for each coin and the corresponding best classifiers, original features, lags and lifts with respect to \cite{mcnally2018}. We see that  \c2p2\ improves the prior work of \cite{mcnally2018} by a margin of 5.1\%-44.1\%. We also implemented \cite{sin2017bitcoin} and tested it on the same training and test split of Bitcoin data. \c2p2 beats \cite{sin2017bitcoin} by 16\% for predicting next-day closing price going up or down. 
    
    Similarly, we see that \c2p2\ improves the prior work \cite{mcnally2018} by a margin of 6.5\%-83\%, 6\%-81.7\% and 6.6\%-65\% respectively on High-High, Low-Low and Open-Open predictions. Our improvements across all coins and tasks are statistically significant with $p < 10^{-5}$ (using the Student's t-test).
    
    

    \begin{table}[tb]
        \centering 
        \begin{tabular}{c c c c c c} 
        \hline\hline 
        Cryptocurrency & Classifier & Features & Lag & AUC & Lift \\ [0.5ex] 
        \hline 
        Binance Coin & L-SVM & E & 21 & 0.717 & 1.347 \\
        Bitcoin & L-SVM & P, E, R & 7 & 0.697 & 1.306 \\
        Bitcoin Cash & LR & E & 10 & 0.761 & 1.406 \\
        Cardano & LR & E & 20 & 0.685 & 1.441 \\ 
        Dash & LR & P & 7 & 0.768 & 1.255 \\
        EOS & LR & R & 5 & 0.662 & 1.395 \\
        Ethereum & NB & R & 6 & 0.709 & 1.263 \\
        Ethereum Classic & RF & E & 12 & 0.681 & 1.406 \\
        IOTA & RF & P, E, R & 29 & 0.709 & 1.371 \\
        Litecoin & L-SVM & P & 12 & 0.705 & 1.159 \\
        Maker & L-SVM & R & 13 & 0.669 & 1.360 \\
        Monero & NB & R & 21 & 0.672 & 1.051 \\
        NEM & L-SVM & P & 8 & 0.719 & 1.193 \\
        NEO & KNN & P, E, R & 3 & 0.671 & 1.312 \\
        Ontology & NB & R & 20 & 0.707 & 1.245 \\
        Ripple & LR & P & 17 & 0.706 & 1.407 \\
        Stellar & RF & P, E, R & 21 & 0.632 & 1.132 \\
        Tether & NB & R & 18 & 0.684 & 1.401 \\
        TRON & KNN & R & 7 & 0.718 & 1.093 \\
        VeChain & LR & P, E, R & 8 & 0.756 & 1.456 \\
        Zcash & RF & P, E, R & 15 & 0.673 & 1.234 \\ [1ex] 
        \hline 
        \end{tabular}
        \caption{Experiment 1 (Close-Close): \C2P2 lift with respect to an LSTM baseline as described in \cite{mcnally2018}. We report the best result for each coin and the corresponding classifier, features, lag, AUC and lift over the baseline. For features, E stands for global economic features, P stands for historical price features, R stands for Reddit-based features. Improvements are statistically significant ($p < 10^{-5}$).} 
        \label{table:baseline_cc} 
    \end{table}

    \subsection{Experiment 2: Contribution of the similarity features in \c2p2}
    We conduct an ablation test by removing similarity components ($\bm{s}_{-c,d}, \forall c \in [1 \ldots C]$ used in Algorithm \ref{alg}) in the task of Close-Close prediction. We report the highest AUCs for all coins in Table \ref{tab:sim_contri} and compute the lifts of \c2p2\ with respect to the AUCs obtained when \c2p2\ does not use similarity features, i.e., the AUC column in Table \ref{table:baseline_cc} divided by the one in Table \ref{tab:sim_contri}.  We observe that the similarity features improve AUCs of all but one coin (Stellar) by a margin of 0.4\% to 17\%. This shows the contribution of similarity features and the power of collectively predicting coins' prices.
    \begin{table}[tb]
        \centering 
        \begin{tabular}{c c c c c} 
        \hline\hline 
       Cryptocurrency & Classifier & Lag & AUC & Lift \\ [0.5ex] 
        \hline 
        Binance Coin & RF  & 8 & 0.657 & 1.092 \\
        Bitcoin & RF  & 25 & 0.628 & 1.110 \\
        Bitcoin Cash & LR & 7 & 0.716 & 1.064 \\
        Cardano & NB & 28 & 0.685 & 1.078 \\
        Dash & KNN  & 5 & 0.726 & 1.058 \\
        EOS & RF  & 10 & 0.613 & 1.080 \\
        Ethereum & KNN  & 16 & 0.658 & 1.078 \\
        Ethereum Classic & LR  & 8 & 0.637 & 1.068 \\
        IOTA & NB  & 20 & 0.638 & 1.111 \\
        Litecoin & RF  & 30 & 0.676 & 1.043 \\
        Maker & L-SVM  & 17 & 0.657 & 1.018 \\
        Monero & RF  & 12 & 0.614 & 1.094 \\
        NEM & RF  & 29 & 0.610 & 1.178 \\
        NEO & KNN  & 12 & 0.617 & 1.088 \\
        Ontology & LR & 2 & 0.672 & 1.052 \\
        Ripple & RF  & 6 & 0.608 & 1.161 \\
        Stellar & RF  & 13 & 0.638 & 0.991 \\
        Tether & L-SVM & 24 & 0.623 & 1.098 \\
        TRON & RF  & 30 & 0.715 & 1.004 \\
        VeChain & LR  & 5 & 0.678 & 1.115 \\
        Zcash & RF  & 15 & 0.645 & 1.044 \\ [1ex] 
        \hline 
        \end{tabular}
        \caption{Experiment 2: \C2P2 lift with respect to \c2p2 without similarity features. We remove similarity features from \c2p2 and report the best result for each coin and the corresponding classifier, lag, AUC, as well as the lift of \c2p2 with respect to this result. Similarity features improve AUCs of all coins except Stellar by 0.4\% to 17.8\%. These improvements are statistically significant ($p<10^{-5}$).}. 
        \label{tab:sim_contri}
    \end{table}
    
\section{Related Work}
    
    Cryptocurrency price prediction in current literature is usually framed as a regression problem, a market simulation to calculate ROI, or as a classification problem in predicting the sign of future price change. Because cryptocurrencies are not managed by a central bank or government, they do not subscribe to the classical economic theories of supply and demand. Instead, additional features, ranging from digital currency specific features to social media trends, are extracted to better predict the price. 
    
    \emph{Bitcoin price prediction} is the topic of most papers in the area of cryptocurrency price prediction. References \cite{sin2017bitcoin, jang2018, h2018rol, mcnally2018} leveraged blockchain features to predict Bitcoin price with varying degrees of success. Sin et al. \cite{sin2017bitcoin} achieved 64\% classification accuracy in predicting the sign of price change using an ensemble of neural networks tuned with genetic algorithms. Jang et al. \cite{jang2018} produced a regression with a Mean Average Percent Error (MAPE) of 0.0138 using Bayesian Neural Networks. McNally et al. \cite{mcnally2018} reported 53\% accuracy in price change sign prediction using an LSTM neural network. Jang et al. \cite{h2018rol} incorporated global economic and currency metrics in addition to blockchain and market data in an LSTM model, obtaining an RMSE of 59.4.  
    
    \emph{Global economic features}, or features which are selected to represent the health and volatility of the global economy, and their relationship to cryptocurrencies were investigated in several works. References \cite{ciaian2015, kristoufek2015, dyhrberg2015} found evidence of effects from macro-financial indices on Bitcoin price. Ciaian et al. \cite{ciaian2015} integrated market forces of supply and demand, investor attractiveness indicators and global macroeconomic variables to explain Bitcoin prices. Dyhrberg \cite{dyhrberg2015} concluded that cryptocurrencies lie somewhere between traditional currencies and commodities. References \cite{h2018rol, catania2018, jang2018} leveraged these financial features in their predictions, as do we in this paper.
    
    \emph{Sentiment Analysis} has been used extensively to understand the impact of public opinions on the price fluctuations of cryptocurrencies. References \cite{kristoufek2013, kristoufek2015, matta2015bitcoin} found evidence of a strong correlation between the magnitude of public interest (using Google and Wikipedia search volumes as an analog) and the price of Bitcoin. Further, \cite{kristoufek2013, garcia2014, kim2016, bukovina2016} provided evidence that the polarity of public opinion is powerful for predicting public interest in Bitcoin. Pant et al. \cite{pant2018recurrent} added sentiment analysis of tweets to an RNN to produce 77.6\% accuracy . Kim et al. \cite{kim2016} achieved an accuracy of over 70\% for Bitcoin, Ethereum and Ripple using sentiment features extracted from online forum discussions.
    
    \emph{Multiple currency prediction} has only been done in a select few papers. Although the literature primarily focuses on Bitcoin, references \cite{smuts2019, kim2016, abraham2018, catania2018} also attempted prediction for other popular cryptocurrencies. Catania et al. \cite{catania2018} performed regression analysis across Bitcoin, Ethereum, Ripple, and Litecoin, and found statistically significant results for them. Using an LSTM model, Smuts \cite{smuts2019} obtained 85\% accuracy for Ethereum using Google Trends features, and 76\% for Bitcoin using Telegram sentiment.
    
    \emph{Collective Classification} models have been widely used in predicting attributes of graph vertices \cite{kong2012meta} and edges \cite{taskar2004link}. Sen et al. \cite{sen2008collective} gave a detailed introduction and experimental comparison of several types of collective classification algorithms.
    They further discussed various heuristics of constructing the features incorporating interdependent information. 
    Differently, \c2p2 builds five kinds of similarities between all pairs of nodes (cryptocurrencies in our case) as relational features. Moreover, none of these models were directly applied to predicting cryptocurrency prices.

\section{Conclusion}
The problem of predicting the up/down movements of cryptocurrencies is of great interest to both the financial industry and to individual consumers. 
We develop the \c2p2\ algorithm that has two innovations: (i) the use of similarities between cryptocurrency feature vectors, and (ii) that uses tentative predictions about $(C-1)$ cryptocurrency's up/down price movements to predict that of the $C$th cryptocurrency. In addition, we use Reddit data for our predictions. We test \c2p2\ on the 21 cryptocurrencies with the highest market capitalization (according to \url{coinmarket.com}) and show that \c2p2\ beats out two recent competitors with substantial lifts which are statistically significant.

\bibliographystyle{IEEEtran}

\bibliography{references}

\end{document}